# Refining adverse drug reaction signals by incorporating interaction variables identified using emergent pattern mining

Jenna M. Reps[a], Uwe Aickelin[a], Richard B. Hubbard[b]

[a]*School of Computer Science, Jubilee Campus, University of Nottingham, NG8 1BB* [b]*School of Medicine, University of Nottingham, Nottingham, NG5 1PB*

**Abstract**

**Purpose:** To develop a framework for identifying and incorporating candidate confounding interaction terms into a regularised cox regression analysis to refine adverse drug reaction signals obtained via longitudinal observational data.

**Methods:** We considered six drug families that are commonly associated with myocardial infarction in observational healthcare data, but where the causal relationship ground truth is known (adverse drug reaction or not). We applied emergent pattern mining to find itemsets of drugs and medical events that are associated with the development of myocardial infarction. These are the candidate confounding interaction terms. We then implemented a cohort study design using regularised cox regression that incorporated and accounted for the candidate confounding interaction terms.

**Results** The methodology was able to account for signals generated due to confounding and a cox regression with elastic net regularisation correctly ranked the drug families known to be true adverse drug reactions above those

*Email address:* jenna.m.reps@gmail.com (Jenna M. Reps )

that are not. This was not the case without the inclusion of the candidate confounding interaction terms, where confounding leads to a non-adverse drug reaction being ranked highest.

**Conclusions** The methodology is efficient, can identify high-order confounding interactions and does not require expert input to specify outcome specific confounders, so it can be applied for any outcome of interest to quickly refine its signals. The proposed method shows excellent potential to overcome some forms of confounding and therefore reduce the false positive rate for signal analysis using longitudinal data.

*Keywords:* Medical Informatics, signal refinement, data mining, observational data, confounding, emergent pattern mining

## 1. Introduction

Negative side effects of medication, termed adverse drug reactions (ADRs), are a serious burden to healthcare **??**. ADRs are estimated as the cause of 6.5% of UK hospitalisations **?** and a study investigating US death due to ADRs reported rates between 0.08-0.12 per 100,000 **?**. Studies have suggested that the rate of ADRs is increasing annually **?**, motivating the improvement of methods for detecting them.

The process of detecting ADRs starts during clinical trials, however clinical trials often lack sufficient power to detect all ADRs for numerous reasons including time limitations, unrealistic conditions and a limited number of people being included **?**. It is then down to post-marketing surveillance to identify the remaining undiscovered ADRs. This involves three stages: signal detection (identifying associations between drugs and outcomes), signal refinement (prioritising/filtering spurious relationships) and signal evaluation (confirming causality after numerous sources of evidence). There has been a big focus towards developing signal detection methods, involving various forms of data such as spontaneous reporting systems **?**, online data **??**, chemical structures **?** and longitudinal observational data **??**. Unfortunately, all the data sources have their own limitations. Spontaneous reporting systems are historically the main source used for post-marketing analysis but often contain missing values, suffer from under- and over-reporting, and rely on people noticing ADRs **?**. Longitudinal observational data have recently been used to complement spontaneous reporting system data for extracting new drug safety information, and are an excellent potential source of information due to the quantity of observational data available and the number of variables recorded. If we could overcome existing issues, mainly confounding, that limit the use of observational data for causal inference then we may be able to aid the discovery of new ADRs.

We are often plagued with confounding when investigating potential causal relationships retrospectively in observational data **?** due to the data collection being non-random. When an association between an exposure and outcome is discovered in observational data, it may often be explained by the presence of confounding. A confounding variable is one that leads to distorted effect estimates between an exposure and outcome due to the confounder being associated with both the exposure and outcome. For a variable to be considered a confounder of an exposure and outcome relationship it must be a risk factor of the outcome, it must be associated with the exposure and it can not lie within the causal pathway between the exposure and outcome.

Consider, for example, the situation where we wish to determine the relationship between a drug given to treat hypertension and myocardial infarction. If we naively look at the incidence of myocardial infarction within a year after treatment for patients given the drug and the incidence of myocardial infarction within a randomly chosen year for patients never given the drug, then we are likely to find that myocardial infarction is more common in those given the drug and conclude that the drug is associated with an increased incidence of myocardial infarction. However, our conclusion is likely explained by confounding, as patients given the drug (those with hypertension) are medically different from those who do not have hypertension. It is likely that some of the patients given the drug have a poor diet or are stressed. Poor diet and stress would have contributed to the hypertension but are also risk factors of myocardial infarction. Therefore poor diet and stress would be confounding factors. To correctly determine a relationship between an exposure and outcome it is important to account for confounding variables. Techniques such as risk adjustment, stratification, or equally distributing the confounding variables between the comparison groups are potential ways to reduce confounding **?**.

Adjusting for confounders in observational data requires identifying the confounders. Although existing methods aim to address confounding, various studies have shown that existing signal generation methods developed for longitudinal observational data have a high false positive rate **??**. This is most likely due to difficulties identifying confounding variables in a data-driven way. Some studies have shown that including a large number of variables, such as drug indications, into drug safety methods can reduce confounding **???**, but none of these methods included interactive terms. A medical illness is likely to be a result of multiple variables interacting. For example, cardiovascular disease is common in patients with a genetic predisposition such as familial hypercholesterolemia and based on lifestyle such as diet and exercise. Therefore, it is interactive terms between medical events or drugs that are most likely to correspond to confounding variables. However, when there are thousands of medical events and drugs, the number of possible interactions is very large. Existing data-driven methods for incorporating interactive terms into regression models include hierarchal lasso, which adds the interactions along with an interaction regularisation term **?**, and methods utilising matrix factorisation **?**. However, these methods are likely to be highly inefficient when there are thousands of variables to consider (which is often the case for observational data). Instead, methods such as emergent pattern mining **?** that can efficient identify outcome specific associations, even when large numbers of variables are being considered, may be more suitable. A similar idea was used to successfully detect survival associate rules **?** based on cox regression and association rule mining. This shows that it is possible to reduce confounding by combining cox regression and association rule mining.

A suitable post-marketing framework that extracts knowledge from longitudinal observational data could be of the form displayed in Figure 1. The first stage of the proposed framework is to apply an efficient large-scale signal generation method to find associations between exposures and outcomes. In the first step the method would efficiently search through all the exposure and outcome possibilities to find associated pairs. An example of a suitable signal generation method is the high dimensionality propensity score (HDPS) **?**. The HDPS works by developing a predictive model for taking the drug and then a matched cohort analysis is applied, where controls are selected based on having a high propensity for taking the drug (the predictive model predicts that they would have the drug). The HDPS can limit confounding by accounting for a large number of variables. Unfortunately, it is not without issues **??** and still often signals many false positives **?**, this highlights the requirement of additional analysis that can reduce the false positive rate. The second step in the framework is the signal refinement, where complex confounding relationships are discovered and incorporated into a more detailed analysis. The output of the signal refinement is a small set of exposure-outcome pairs that are prioritized for signal evaluation. The final step would be to formally evaluate the remaining signals using a number of different data sources, as establishing a causal relationship requires an accumulation of evidence.

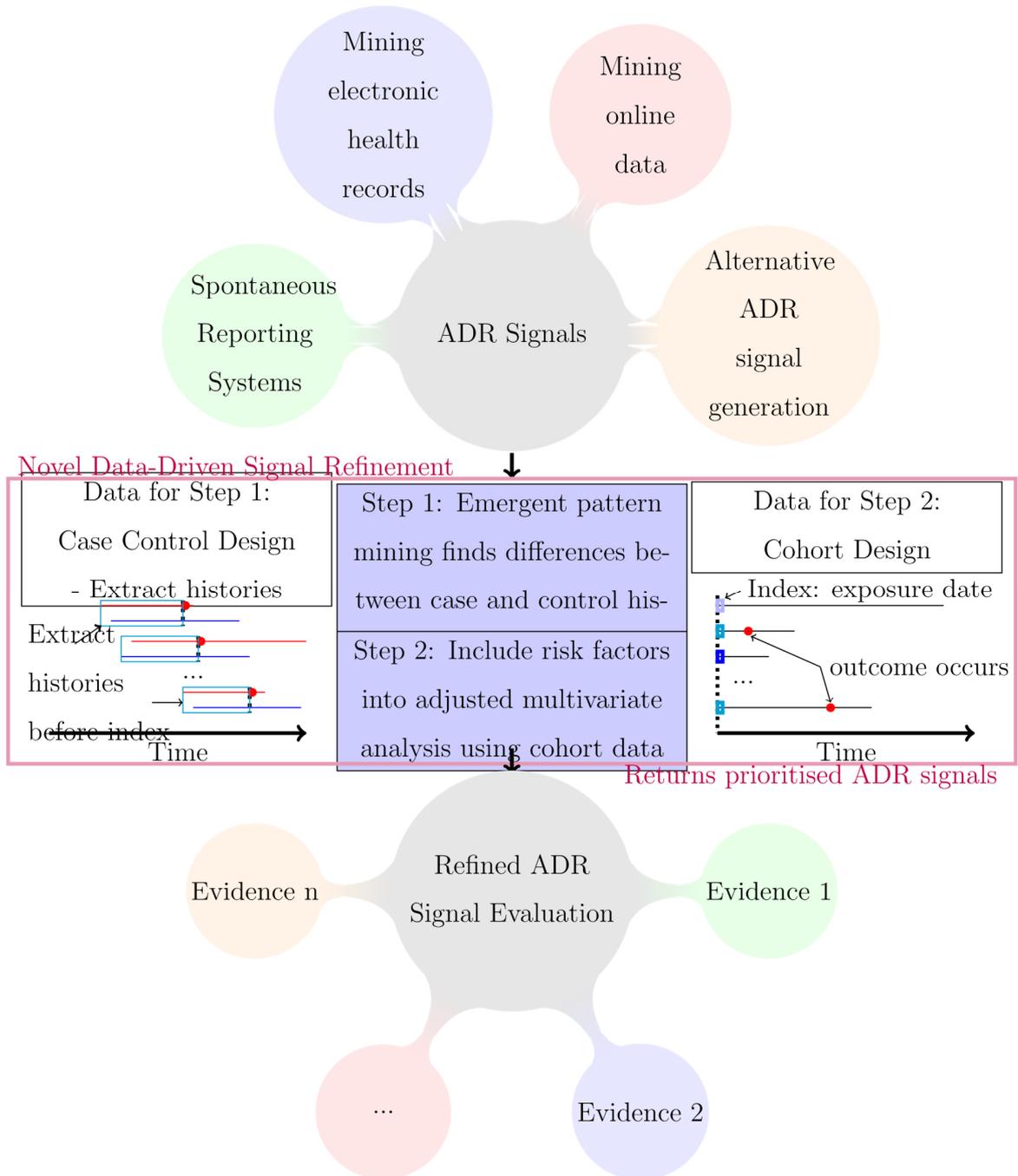

Figure 1: Overview of the methodology. Our proposed data-driven signal refinement is applied to ADR signals generated elsewhere to prioritize/filter them before formal evaluation. It will filter some signals that can be explained by confounding and prioritizes the remaining signals ready for evaluation.

In this paper we focus on the signal refinement stage, as there are no data-driven methods to refine signals, but numerous signal generation and evaluation methods exist. The objective of this research is to develop a data-driven signal refinement methodology that can be applied after ADR signal generation using longitudinal observational data to filter and re-rank the signals by addressing complex confounding. We will test the data-driven methodology by analysing the relationship between numerous drugs and the outcome myocardial infarction (MI). We are exploring three goals:

1. Whether emergent pattern mining can be used to identify candidate interaction confounding covariates in a data-driven way.
2. Whether the inclusion of interaction confounding covariates into a regression analysis can reduce confounding and be used for data-driven ADR signal refinement.

3. Whether lasso and ridge regularisation are suitable techniques to enable the inclusion of a large number of potential interaction covariates.

## 2. Materials and Methods

*2.1. Materials*

The longitudinal observational database used in this study is The Health Improvement Network (THIN) database (www.thin-uk.com). THIN contains complete medical records for patients registered at a participating general practice within the UK. At present approximately 6% of the UK general practices are participating, resulting in THIN containing data on over 4 million active patients. The validity of the THIN database for pharmacoepidemiology studies has been investigated **?** and it was shown that the data appear to be representative of the UK population.

THIN contains time-stamped entries of drugs that are prescribed and medical events that the general practices are made aware of. Prescribed drugs are recorded with a British National Formulary (BNF) code indicating the family of the drug prescribed. Medical events are recorded using the Read coding system. The Read codes used in this study to identify myocardial infarction (MI) are available in Supplement A.

The drug families (represented by BNF codes) investigated in this study are presented in Table 1 along with the ground truth (the known relationship between each drug family and MI), the number of prescriptions eligible for inclusion in the study and number of MI that occurred after an eligible prescription of each BNF. The ground truth relationship between these BNFs and MI are as specified in the Health Outcome of Interest (HOI) reference set used by the Observational Medical Outcomes Partnership (OMOP) **?**.

Table 1: Summary of number of patients prescribed each drug family and number of patients who experience MI within 5 years of their first time prescription.

| Drug Family | BNF[+] | No. First time prescriptions | No. of MI* after BNF | MI is a known ADR? |
|---|---|---|---|---|
| Typical antipsychotics | 04020100 | 10061 | 44 | Yes |
| Benzodiazepines | 04010200 | 74582 | 277 | No |
| Tricyclic antidepressants | 04030100 | 61384 | 221 | Yes |
| Antibiotics 1 | 05010300 | 78296 | 247 | No |
| Antibiotics 2 | 05010800 | 91515 | 258 | No |
| Bisphosphonates | 06060200 | 8967 | 67 | No |

Note: [+] BNF: British National Formula, * MI: Myocardial Infarction

## 2.2. Analysis Methods

### 2.2.1. Emergent Pattern Mining

Emergent pattern mining is a type of association rule mining that aims to find differences between databases. Formally, let $I = \{i_1, i_2, ,i_n\}$ be the set of $n$ items and $t = \subset X$ be a transaction containing a set of items. A database is a collection of transactions, denoted by $D = \{t_1, t_2, ,t_m\}$. The support of an itemset $X$ in database $D$ is the proportion of the database transactions that contain $X$,

$$Supp_D(X) = |\{t_i \in D | X \subset t_i\}|/|D| \qquad (1)$$

where $|D|$ is the number of transaction in the database. An itemset is frequent if it has a support greater than some user defined minimum support value.

In the drug safety context, *I* represents the observational databases clinical vocabulary (e.g., International Classification of Diseases and RxNorm), and a patients transaction may correspond to the set of codes that they have recorded in the database during a specified time interval. An itemset is a collection of clinical vocabulary codes (e.g., {drug1, illness350, drug924}), and the support of the itemset is the fraction of the database's patient transactions that contain the itemset.

The idea of emergent pattern mining is to find itemsets that have a higher support in one database than others **?**. A simple method for determining the emergent patterns is to find the itemsets that have a sufficiently high support in one of the databases and then calculate the lift,

$$lift(X) = Supp_{D1}(X)/Supp_{D2}(X) \qquad (2)$$

However, this measure is vulnerable to volatility due to small support values and a bias-adjusted lift may be more suitable,

$$biaslift(X) = (Supp_{D1}(X) + 1)/(Supp_{D2}(X) + 1) \qquad (3)$$

this measure adds a bias towards 1 when the support is low and is a common approach for dealing with rare events in drug safety[28].

*2.2.1.1 Cox Regression*

Cox regression was chosen as this enables the consideration of the time to event (enabling long term effects to be evaluated) and the inclusion of rightcensored data. Using a method that ignores the time-to-event, such as logistic regression, would likely result in bias, as patients who leave the practice or die from alternative illnesses (right-censored patients) would not be included into the analysis **?**. The cox regression model assumes that the hazard at time *t* given the covariates x and baseline hazard is,

$$h(t|\mathbf{x}) = h_0(t) exp(\sum_j \mathbf{x_j}\beta_j) \qquad (4)$$

The parameter *β* is determined by maximising the partial likelihood **?**.

### 2.2.1.2 Regularised Cox Regression

Regularisation is a technique to prevent models overfitting by adding a constraint to the complexity of the model. This is necessary when there are too many covariates in the model. In terms of cox regression, regularisation techniques add a constraint to the total size of the coefficients in the model. There are two main forms of regularisation penalty functions, ridge **?** and lasso **?**. The elastic net regularisation combines both the lasso and ridge penalty functions **?**. Lasso regression is a regularised regression that adds a constraint to the total sum of the absolute value of the model coefficients into the optimisation,

$$\lambda ||\beta||_1 = \lambda \sum_j |\beta_j| \qquad (5)$$

This effectively causes some of the coefficients to disappear and is useful when there are a smaller number of medium or large effects. It has problems when there are correlations between covariates, as it will only pick one of the correlated covariates coefficients to be non-zero. Lasso regression has been used successfully with the aim of overcoming confounding in drug safety **?** but no existing approach has included interactive terms.

Ridge regression adds a constraint to the sum of the coefficients squared into the optimisation,

$$\lambda||\beta||^2 = \lambda \sum_j \beta_j^2 \qquad (6)$$

this causes all the coefficients to shrink to zero as lambda tends to infinity. Ridge regression is useful if there are small effects, which may be the case for ADRs. Finally, the elastic net linearly combines both ridge and lasso regression penalties,

$$\lambda_1||\beta||_1 + \lambda_2||\beta||_2 = \lambda_1 \sum_j |\beta_j| + \lambda_2 \sum_j \beta_j^2 \qquad (7)$$

using the parameter $\alpha \in [0,1] = \lambda_1/(\lambda_1 + \lambda_2)$.

The regularised cox regression can be used to indicate the relative importance of each variable in determining the time-to-event of MI. If the drug suspected of causing MI is deemed to have a large coefficient by the regularised cox regression then this strengthens our suspicion, especially when alternative risk factors for developing MI are accounted for within the model.

### 2.3. Method

The proposed signal refinement methodology has two steps. Step 1 involves applying emergent pattern mining to identify sets of medical conditions and medication that are risk factors for developing the outcome (candidate positive confounder interactions) or factors that reduce the likelihood of developing the outcome (candidate negative confounder interactions). Step 2 involves implementing a cohort study design and a risk adjusted multivariate analysis (e.g. regularised cox regression) with risk adjustment terms accounting for the candidate confounder variables identified in step 1.

### 2.3.1. Step 1 Candidate interactive confounder identification

#### 2.3.1.1 Select Data

We imitated the data extraction of a case control study design to extract data for the emergent pattern mining. In this step we did not apply a case control design to investigate the effect of exposure, we just extracted the case control study design data for the emergent pattern mining. The cases are the patients who have the outcome recorded; their index date is the first time the outcome is recorded. Then two controls are selected for each case by matching on age (plus or minus 1 year), gender and general practice and finding the two patients that have the closest registration date to the case. We chose to match two controls per case as the majority of cases had two potential matches and the greater the number of controls, the more likely each itemsets support value will approximate the actual prevalence. Each control's index date is the date of their matching cases index date. The exclusion criteria are: patients with a history of any of the exposures being investigated prior to index were excluded. Controls must have zero recordings of the outcome in the data. The inclusion criteria are: patients that are aged between 18 and 70 and patients must have been registered for at least a year prior to their index date. Figure 2 illustrates the implemented case control style design for extracting suitable data to apply emergent pattern mining to. The emergent pattern mining will find sets of drugs/medical conditions

Table 2: Example of transactions found in *D*1 (the patients with the outcome recorded in the database)

| Patient ID | Index Date (date of outcome) | General Practice | Sex | Date of Birth | Medical transactions* |
|---|---|---|---|---|---|
| 1 | 01-02-2008 | 2 | M | 02-07-1967 | Drug1, Drug110, Illness1020, Illness15001 |
| 15 | 23-01-2001 | 20 | M | 16-04-1936 | Drug204, Drug110, Illness1020, Illness3 |
| 37 | 12-12-2001 | 1 | F | 03-12-1942 | Illness4 |
| … | … | … | … | … | … |

Note: * - All recorded prescriptions/medical items prior to index date

that occur more often prior to index for the cases (those with the outcome, in our example MI) compared to the controls. These correspond to potential risk factors for the outcome.

### 2.3.1.2 Create patients transaction baskets

We then create two databases $D1$ and $D2$, see Tables 2-3. $D1$ contains the case patients medical transactions, i.e. the set of all medical events or drugs recorded prior to the index date for each patient in the case control study. $D2$ contains the control patients medical transactions, i.e. the set of all medical conditions or drugs recorded prior to the index date.

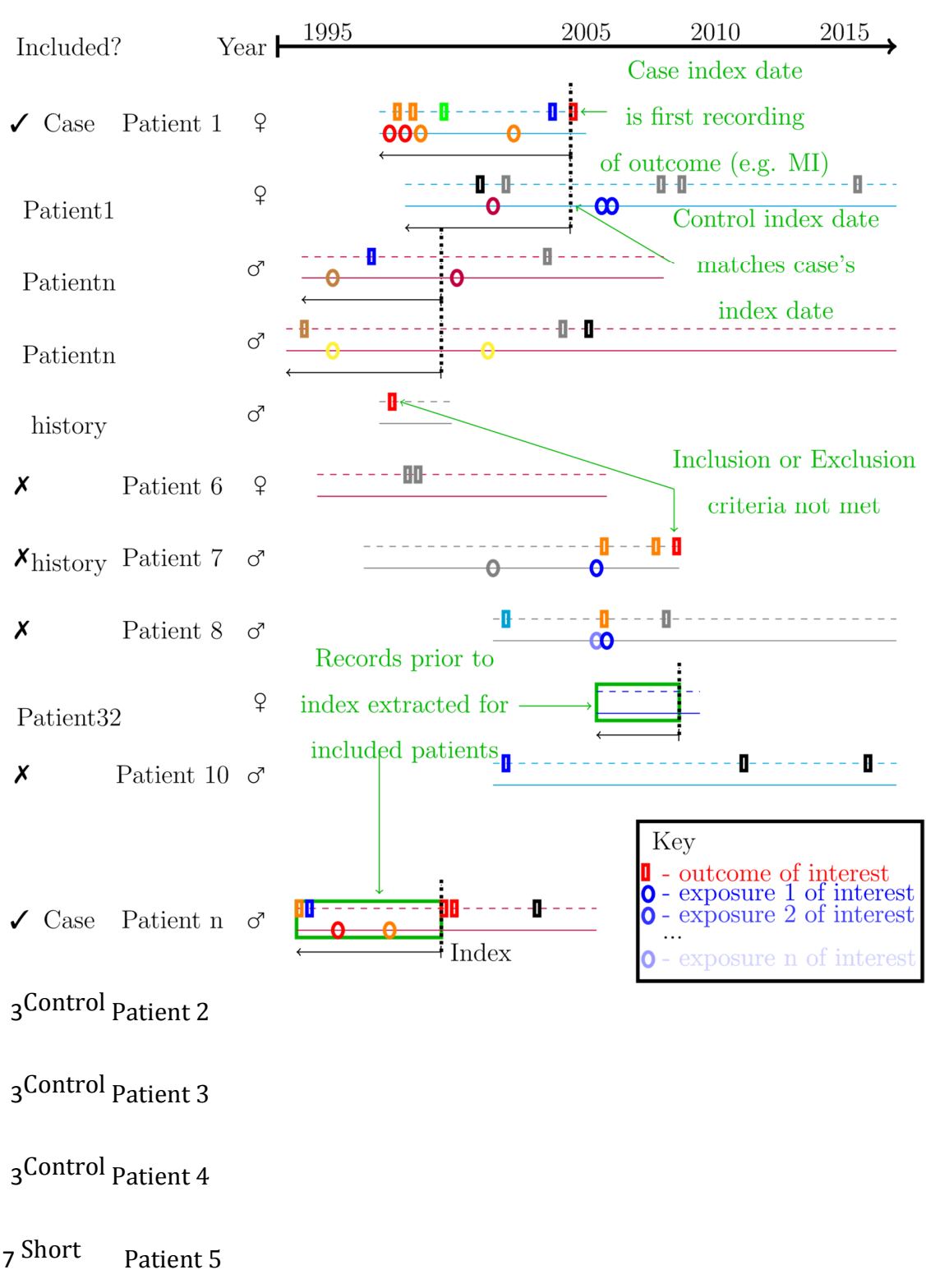

Exposure

3Control Patient 9

...

Figure 2: Step 1 study design illustration: For each patient in the longitudinal observational database we have their prescription timeline (dashed lines) and condition/observation timeline (straight line) since registration until they leave the practice or die. Circles represent recorded conditions and rectangles represent recorded medication. Each colour of the shape represents a type of drug or condition, for example a red rectangle represents MI. The patient's timeline colour corresponds to their general prac-15 tice. The index date used for step 1, the first outcome record date, is represented by a dotted line.

Table 3: Example of transactions found in $D2$ (the controls matched on practice, gender and age)

| Patient ID | Index Date | General Practice | Sex | Date of Birth | Medical transactions* |
|---|---|---|---|---|---|
| 1020 | 01-02-2008 | 2 | M | 23-09-1967 | Drug1, illness43, Illness2, Drug203 |
| 1031 | 01-02-2008 | 2 | M | 01-05-1968 | Illness3 |
| 43 | 23-01-2001 | 20 | M | 08-02-1936 | Illness3, Illness4010 |
| ... | ... | ... | ... | ... | ... |

Note: * - All recorded prescriptions/medical items prior to index date

### 2.3.1.3 Identify emergent patterns

We then apply frequent itemset mining to *D*1 with a minimum support of 0.001 to find all the cases' frequent itemsets $I_{D1}$ and apply frequent itemset mining to *D*2 with a minimum support of 0.0005 to find all the controls' frequent itemsets $I_{D2}$. The positive measure of interestingness is calculated by dividing the support of each itemset, $X \in I_{D1}$, by its support in *D*2 but we add a bias to reduce the measure to one for small support values. For each $X \in I_{D1}$ we calculate,

$$biaslift_{D2}^{D1}(X) = \begin{cases} \left(\frac{Supp_{D1}(X)+1}{Supp_{D2}(X)+1}\right) & \text{if } X \in I_{D1} \cap I_{D2} \\ (Supp_{D1}(X) + 1) & \text{if } X \in I_{D1}, X \notin I_{D2} \\ 0 & \text{otherwise} \end{cases} \quad (8)$$

The candidate positive confounding variables are the $\{X \in I_{D1} | biaslift_{D2}^{D1}(X) > 1\}$. The negative emergent patterns are found in a similar manner but we apply frequent itemset mining to *D*2 with a minimum support of 0.001 to find all the controls frequent itemsets $I_{D2}$ and apply frequent itemset mining to *D*1 with a minimum support of 0.0005 to find all the cases frequent itemsets $I_{D1}$. We then calculate $biaslift_{D1}^{D2}(X)$ and determine the candidate negative confounding variables as $\{X \in I_{D2} | biaslift_{D1}^{D2}(X) > 1\}$.

For example, if we found that the itemset {*Drug*110,*Illness*1020} had a support of 0.05 in *D*1 (the itemset is found in the medical transactions for every 5 out of 100 patients with MI) and a support of 0.005 in *D*2 (the itemset is found in the medical transactions for every 5 out of 1000 control patients). Then $biaslist_{D2}^{D1}(\{Drug110, Illness1020\})$ = 1.05/1.005 = 1.045 and would be considered a candidate positive confounding variable.

The minimum support value was a trade off between finding as many itemsets as possible and finding useful itemsets. The minimum support of 0.001 means any identified itemset would be found in a minimum of 1 in 1000 patients with MI. A support value less than 0.001 would likely find irrelevant itemsets, as they would be too rare to be useful. The support was reduced for the comparison database to increase the chance that the itemset is found in both databases frequent itemsets.

Although it is possible to identify itemsets of any size using frequent itemset mining, in this paper we added a constraint to only find itemsets containing 5 items or less. This ensured the method was highly efficient and in general we found that itemsets containing more than 5 items often had small support values, so this constraint is unlikely to affect the results. Overall we were efficiently able to identify interactions containing up to 5 variables that are associated with developing MI. We found 23808 variables recorded into THIN for patients who had MI recorded, this means we searched through $6.35 \times 10^{14}$, $\sum_{(k=2)}^{5} \binom{23808}{k}$, possible interactions to find the most common ones that are associated with developing myocardial infarction.

Table 4: Example of the results of the emergent pattern mining. The most suitable variables are selected based on them having the greatest *biaslift* value.

| Itemset | D1 Support | D2 Support | $biaslift_{D2}^{D1}$ | $biaslift_{D1}^{D2}$ |
|---|---|---|---|---|
| {Drug110,Illness1020} | 0.05 | 0.005 | 1.045 | 0.957 |
| {Illness3, Illness4010} | ¡0.0005 | 0.001 | - | 1.001 |
| {Drug201, Drug 104, Illness1023} | 0.0012 | ¡0.0005 | 1.0012 | - |
| ... | ... | ... | ... | ... |

*2.3.1.4   Select emergent patterns*

Table 4 displays the results of the emergent pattern mining approach (a list of sets of medical items that occur more for patients with/without the outcome). Finally we select the top *k* candidate positive confounding variable itemsets (with the greatest $biaslift_{D2}^{D1}$) and the top *k* negative confounding variable itemsets (with the greatest $biaslift_{D1}^{D2}$). These will be included into the cox regression using indicator variables to mark their presence or absence in each cohort patients medical history prior to their cohort index date. In this study we chose *k* = 200 as this seemed to give a sufficient number of candidate confounders while still enabling the methodology to be implemented
efficiently.

*2.3.1.5   Justification*

The motivation for applying a case control study combined with emergent pattern mining is that we can find illnesses/drugs that are more common in the patients who have the exposure compared to those who never develop the exposure. These are the potential risk factors for the exposure, the first criterion necessary for a variable to be a confounder to the exposure-outcome relationship. By excluding patients from the case control data extraction who have the exposures being analysed, we are preventing the discovery of risk factors that are within the causal pathway between exposure and outcome (as this would violate the third criterion necessary for a variable to be a confounder).

*2.3.2. Step 2 Cohort study with risk adjusted multivariate analysis*

*2.3.2.1 Select Data*

We implement a cohort study design with a 5-year follow up period. For each drug family of interest being investigated, we find all the patients prescribed the drug family for the first time between the years 2005 and 2010 and set their index date as the first time prescription of the drug family. The inclusion criteria are: patients are aged between 18 and 65 at index (because we use a 5 year follow up and the emergent patter mining was for patients aged 70 or younger) that have more than 1 years history prior to the index date. Figure 3 illustrates the study design implemented in step 2, where each shade of blue circle corresponds to one of the exposures being investigated. If a patient was prescribed more than one of the exposures being investigated for the first time during 2005 and 2010, then they would be included multiple

Table 5: Example of the results of the emergent pattern mining. The most suitable variables are selected based on them having the greatest *biaslift* value.

| Patient | Age | Sex | Exposure$_1$ (06060200) | ... | Exposure$_N$ (04010200) | Itemset$_1$ | ... | Itemset$_{400}$ | Outcome |
|---|---|---|---|---|---|---|---|---|---|
| 2032 | 64 | M | 1 | ... | 1 | 0 | ... | 1 | 1 |
| 10570 | 49 | F | 0 | ... | 0 | 0 | ... | 0 | 0 |
| 34 | 58 | M | 1 | ... | 1 | 0 | ... | 0 | 1 |
| ... | ... | ... | ... | ... | ... | ... | ... | ... | ... |

times into the cohort but with different index dates, see patient 8 in Figure 3.

The covariates used in the cox regression model are the patients age/gender, indicator variables for each of the drug families being investigated (1 if the patient had a prescription of the drug family recorded prior to or on the index date and 0 otherwise) and indicator variables for each of the 400 candidate confounder itemsets identified in step 1 (1 if the patient has all items in the itemset recorded prior to index date and 0 otherwise). An example of the data used to learn the cox regression model is displayed in Table 5.

*2.3.2.2   Cox Regression Models*

We then implement various regularised cox regression models (using the elastic net regularisation detailed in equation 7 with various $\alpha \in [0,1]$ values including lasso:$\alpha = 1$ and ridge:$\alpha = 0$) using age/sex, drug family history indicators and candidate interaction confounders covariates. For comparison we also implement a standard cox regression model that only used the tion. Each colour of the shape represents a type of drug or condition, for example a red rectangle represents MI and the different shades of blue circles represents the different exposures of interest. The patient's timeline colour corresponds to their general practice and their gender is represented by gender symbols.

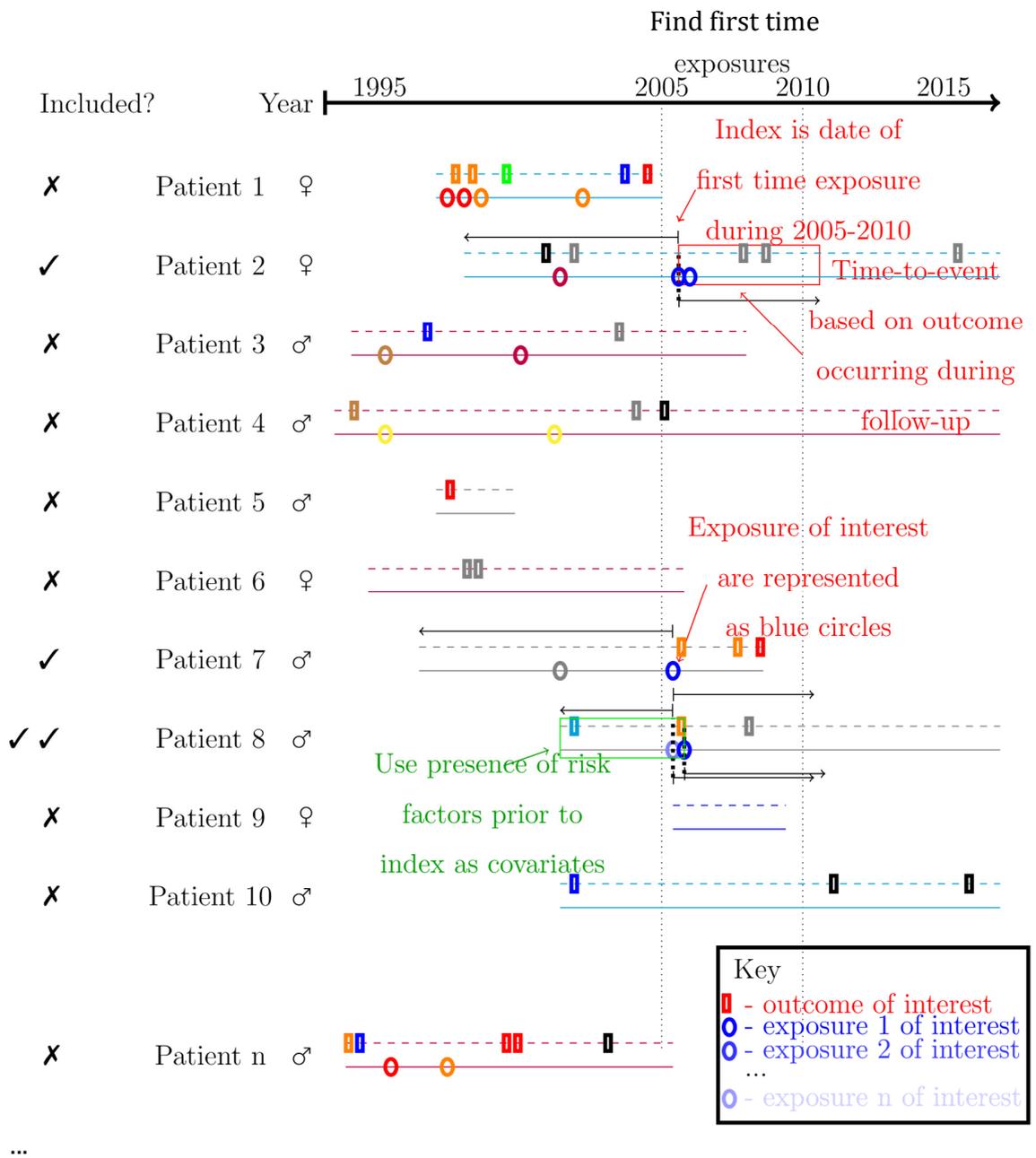

Figure 3: Step 2 study design illustration: For each patient in the longitudinal observational database we have their prescription timeline (dashed lines) and condition/observation timeline (straight line) since registration until they leave the practice or die. Circles represent recorded conditions and rectangles represent recorded medica21

age/sex and drug family indicator covariates.

*2.3.2.3 Validation*

Following standard computer science methodology we split the data into test/train sets. 50% of the data were used to train the cox regression models, as this was ample, and the remaining 50% of the data were used to validate the models. The summary time-dependant area under the receiver operating characteristic curve (AUC) **?** was calculated to validate the model fit.

*2.3.2.4 Software*

The data were stored and manipulated in MS SQL Server and the analysis was conducted using the open source analysis language R **?**. The frequent pattern mining was implemented using the arules package's Apriori algorithm **?**, the standard cox regression was implemented using the survival package **?**, the regularised cox regression was implemented using the glmnet package **?** and the model validation was implemented using the survAUC package.

## 3. Results

*3.1. Emergent pattern mining results*

We identified 77,246 eligible patients who had MI recorded and these patients transactions were included into $D1$. We matched 150,304 patients into $D2$. The frequent pattern mining was applied to 23,808 items in D1 and found 3,886,408 frequent itemsets for $D1$ with a minimum support of 0.001 (9920792 with a minimum support 0.0005). For $D2$ there were 26,705 items, with association rule mining identifying 2,092,949 frequent itemsets for $D2$ with a minimum support of 0.001 (5,502,600 with a minimum support 0.0005). In total we identified 3,838,643 potential candidate positive confounders and 57,507 potential negative confounders, however only the top 200 positive and top 200 negative candidate confounders were selected.

One interesting observation of the emergent pattern mining results is that the there were more frequent itemsets discovered for the patients with MI compared to the control group in step 1. One explanation of this is that patients with MI are likely to be similar and share the common risk factors or MI preceding events, so the $D1$ dataset will be less sparse compared to $D2$, which consists of a more varied population. Another reason $D2$ is sparser is that it contained healthier (non-MI) patients, some of which may not have many items recorded. In future work it would be interesting to restrict $D2$ to patients with certain criteria, such as having a minimum number of items, and investigating how this effects the emergent patterns that are discovered.

*3.2. Cox regression results*

In total we used data for 162,889 patients for the cox regression analysis. The model concordance for the standard cox regression was 0.82 and none of the covariates violated the proportional hazard assumption. Table 6 presents the ranking of the drug families based on the coefficient values obtained from a cox regression that only used the age/sex and drug family indicator covariates. The non-ADR bisphosphonates (BNF 06060200) was found to have the greatest hazard ratio.

Table 7 presents the results of the elastic net regularised cox regression with the 6 drug families, age/sex and 400 candidate interactive confounders as covariates. The two known ADRs were correctly ranked above the non-

Table 6: Results of cox regression that does not include the candidate confounding interaction variables.

| Rank | Drug Family Covariate | Coefficient | Hazard Ratio (95% CI) | Truth |
|---|---|---|---|---|
| 1 | 06060200 | 0.563 | 1.757 (1.326-2.326) | Non-ADR |
| 2 | 04020100 | 0.398 | 1.488 (1.103-2.009) | ADR |
| 3 | 04030100 | 0.255 | 1.291 (1.086-1.534) | ADR |
| 4 | 05010300 | 0.150 | 1.162 (0.977-1.382) | Non-ADR |
| 5 | 05010800 | 0.136 | 1.145 (0.958-1.369) | Non-ADR |
| 6 | 04010200 | 0.069 | 1.071 (0.896-1.280) | Non-ADR |

ADRs when candidate confounders were incorporated via a ridge cox regression.

On the 50% testing data the time dependant AUC for the standard cox regression was 0.816, the lasso regularised cox regression obtained an AUC of 0.825 and the ridge regularised cox regression obtained an AUC of 0.785, indicating strong model fits.

## 4. Discussion

This is the first methodology proposed for incorporating candidate interaction confounder covariates into a cox regression for drug safety. The standard cox regression that only considered indication of the various drug families on the day of or prior to index, age and sex ranked bisphosphonates (BNF 06060200), a non-ADR, as the most likely to cause MI. However, incorporating the candidate interaction confounders into the elastic net regression with small values for $\alpha$ reduced the confounding in the analysis

Table 7: The results of the elastic net cox regression. No. $\beta$s is the number of variables selected by the model.

| | Parameters | | | | | |
|---|---|---|---|---|---|---|
| $\alpha$ | 1 (Lasso) | 0.6 | 0.3 | 0.1 | 0.05 | 0 (Ridge) |
| $\lambda^*$ | $3.7\times10^{-4}$ | $1.4\times10^{-4}$ | $3.3\times10^{-4}$ | $8.3\times10^{-4}$ | $1.4\times10^{-3}$ | $8.1\times10^{-3}$ |
| No. $\beta$s | 6 | 26 | 30 | 51 | 73 | 266 |

—

| Covariate | Coefficient values $\beta$ (Rank) | | | | | | Truth |
|---|---|---|---|---|---|---|---|
| 04030100 | 0 | 0.030 (17) | 0.006 (23) | 0.011 (33) | **0.0123** (38) | **0.0112** (64) | ADR |
| 04020100 | 0 | 0 | 0 | 0.003 (38) | 0.011 (41) | 0.011 (67) | ADR |
| 06060200 | 0 | **0.12** (6) | **0.0152** (21) | **0.012** (30) | 0.0120 (39) | 0.009 (80) | Non-ADR |
| 05010300 | 0 | 0 | 0 | 0 | 0 | 0.008 (84) | Non-ADR |
| 04010200 | 0 | 0 | 0 | 0 | 0 | $-7\times10^{-4}$ (166) | Non-ADR |
| 05010800 | 0 | 0 | 0 | -0.006 (42) | -0.022 (66) | -0.018 (262) | Non-ADR |

Note:

-$\lambda^*$ is the maximum lambda that was within 1 standard deviation of the minimal cross-validation error as this was a trade off between maximising regularisation and minimising error.

- Bold entries correspond to the top ranked drug family

and correctly ranked the two known ADRs, tricyclic antidepressants (BNF 04030100) and typical antipsychotics (BNF 04020100), above bisphosphonates. As $\alpha$ increased from 0 in the elastic net regression, the framework was unable to reduce the confounding completely and still incorrectly ranked bisphosphonates as the most likely cause of MI. As $\alpha$ increased towards 1 (lasso regression), it was generally observed that the regularised regression did not identify any of the drug families as causes of MI, however, at certain small values of $\alpha$ (e.g., $\alpha = 0.1$, $\alpha = 0.05$), the lasso effect nicely filters some of the non-MI causing drug families, although bisphosphonates were still ranked above some of the known ADRs. This shows that elastic net cox regression including the candidate interactive confounders has the potential to not only re-rank ADRs signals, but also filter the false positive signals. The results suggest the ideal ADR signal refinement framework should take a multiple step approach and combine results obtained with various values of $\alpha$. For example, use elastic net cox regression with a small $\alpha$ to filter signals firstly and then use elastic net cox regression with $\alpha$ set to zero to rank the remaining signals.

One possible reason why the elastic net cox regression with $\alpha = 0$ was able to correctly rank the drug families is that it incorporated more covariates into the model, even those with small effects. The result that the elastic net cox regression with a small $\alpha$ was better than lasso regression, also provides some evidence that the candidate interactive confounder covariates are appropriate, as including them into the cox regression reduced confounding. The fact that lasso regression shrunk all the drug families coefficients to zero is not unexpected, as lasso regression identifies large/medium effects, but ADRs that are difficult to distinguish from confounding are likely to have a small effect size. This suggests that the choice of $\alpha$ will depend on the outcome being investigated (e.g., is it common/rare) and the relationship between the drugs and outcome. Rather than relying on the results obtained by picking the optimal lambda for lasso regression, it may be suitable to investigate the results returned for a range of lambdas and identify the order in which the drug family variables are entered into the regularised models as the regularisation decreases. The last *n* occurring drug family variables (or those with negative coefficients) could then be filtered out. Another possible explanation for elastic net cox regression with $\alpha = 0$ outperforming lasso is due to unmeasured confounding. If there are many unmeasured confounders, then adding more candidate interaction confounder covariates may increase the chance of identifying a proxy variable for some unmeasured confounding. Therefore, as elastic net cox regression with a small $\alpha$ includes more covariates it will have an advantage over lasso regression, as it is may overcome some unmeasured confounding.

We purposely chose MI in this study as the outcome of interest due to it having a high background rate, so confounding is a common issue, and also due to missing data problems being common (MI may only be recorded in secondary care and missed from the primary care record). For example, some patients experiencing a MI may not have the event recorded in their primary care records as they may go to hospital. Hospitalisation may also cause the recording date to be incorrect in THIN. By purposely choosing a difficult outcome, we are testing whether the proposed method can overcome both confounding and data recording issues. However, the proposed method does require that the patients have medical histories; otherwise it would not be possible to identify confounders related to outcome progression. This may limit which databases the methodology can be applied to, as longitudinal observational data with short observation periods are less suitable, although even a short period of history may be useful for reducing confounding. Therefore, it would be useful in future research to determine whether the proposed method can successfully reduce confounding when applied to other outcomes or when implemented for signal refinement using other longitudinal observational data.

Overall the methodology shows promise at refining ADRs by reducing confounding. The results show that combining emergent pattern mining and elastic net cox regression is an effective and efficient data-driven framework for reducing the issue of confounding that is common in longitudinal observational studies. However, further research is required to identify:

- The optimal value for $k$ (the number of candidate confounder covariates)

- The most suitable choice of $\alpha$ for the elastic net (or combination)

- The types of outcomes that are suitable for the method (e.g., common, rare, progressive, chronic, acute)

The proposed methodology has the advantage of being efficient as it is datadriven, it is applicable for analysing both short-term and long-term ADRs and it can remove false ADR signals. However, formal epidemiology study designs will still need to be applied to the unfiltered signals before causality can be established.

## 5. Conclusions

In this paper we proposed a novel framework to efficiently enable the inclusion of high-order interactive terms, potentially representing confounders, into a cox regression analysis to refine ADR signals. The framework combines emergent pattern mining, that searches billions of possible interactions to identify terms potentially corresponding to confounders, and regularised cox regression. We investigated the framework by applying it to investigate how likely six different drug families are to cause

MI. The drug families were chosen as they are from a reference set developed to evaluate ADR signal detection methods and there have been numerous studies investigating their relationships with MI. A standard cox regression only considering the risk adjustments of age, sex and indicators for the five drug families was shown to be negatively effected by confounding. However, our proposed automated framework for incorporating candidate confounder interaction terms into an elastic net cox regression was shown to be more resilient to confounding and correctly ranked the drug families in order of how likely they are to cause MI. This shows excellent potential for reducing the current high false positive rate issue when using longitudinal observational data for drug safety.

In future work the framework's robustness should be tested by implementing the framework to refine ADR signals for alternative outcomes or by using alternative longitudinal observational data.

## 6. References

E. C. Davies, C. F. Green, S. Taylor, P. R. Williamson, D. R. Mottram, M. Pirmohamed, Adverse drug reactions in hospital in-patients: a prospective analysis of 3695 patient-episodes, PLoS one 4 (2) (2009) e4439. doi:10.1371/journal.pone.0004439.

M. Pirmohamed, S. James, S. Meakin, C. Green, A. K. Scott, T. J. Walley, K. Farrar, B. K. Park, A. M. Breckenridge, Adverse drug reactions as cause of admission to hospital: prospective analysis of 18 820 patients, BMJ 329 (7456) (2004) 15–19. doi:10.1136/bmj.329.7456.15.

G. Shepherd, P. Mohorn, K. Yacoub, D. W. May, Adverse drug reaction deaths reported in United States vital statistics, 1999–2006, Ann. Pharmacother. 46 (2) (2012) 169–175. doi:10.1345/aph.1P592.

T.-Y. Wu, M.-H. Jen, A. Bottle, M. Molokhia, P. Aylin, D. Bell, A. Majeed, Ten-year trends in hospital admissions for adverse drug reactions in England 1999–2009, J. R. Soc. Med. 103 (6) (2010) 239–250. doi: 10.1258/jrsm.2010.100113.

L. Ha¨rmark, A. Van Grootheest, Pharmacovigilance: methods, recent developments and future perspectives, Eur. J. Clin. Pharmacol. 64 (8) (2008) 743–752. doi:10.1007/s00228-008-0475-9.

E. P. van Puijenbroek, A. Bate, H. G. Leufkens, M. Lindquist, R. Orre, A. C. Egberts, A comparison of measures of disproportionality for signal detection in spontaneous reporting systems for adverse drug reactions, Pharmacoepidemiol. Drug Saf. 11 (1) (2002) 3–10. doi:10.1002/pds.668.

J. Bian, U. Topaloglu, F. Yu, Towards large-scale twitter mining for drugrelated adverse events, in: Proceedings of the 2012 international workshop on Smart health and wellbeing, ACM, 2012, pp. 25–32. doi:10.1145/ 2389707.2389713.

J. M. Reps, J. M. Garibaldi, U. Aickelin, D. Soria, J. E. Gibson, R. B. Hubbard, A novel semisupervised algorithm for rare prescription side effect discovery, IEEE J. Biomed. Health. Inform. 18 (2) (2014) 537–547. doi: 10.1109/JBHI.2013.2281505.

J. Scheiber, J. L. Jenkins, S. C. K. Sukuru, A. Bender, D. Mikhailov, M. Milik, K. Azzaoui, S. Whitebread, J. Hamon, L. Urban, et al., Mapping adverse drug32

reactions in chemical space, J. Med. Chem. 52 (9) (2009) 3103–3107. doi:10.1021/jm801546k.

M. J. Schuemie, R. Gini, P. M. Coloma, H. Straatman, R. M. Herings, L. Pedersen, F. Innocenti, G. Mazzaglia, G. Picelli, J. van der Lei, et al., Replication of the OMOP experiment in Europe: evaluating methods for risk identification in electronic health record databases, Drug Saf. 36 (1) (2013) 159–169. doi:10.1007/s40264-013-0109-8.

J. M. Reps, J. M. Garibaldi, U. Aickelin, D. Soria, J. E. Gibson, R. B. Hubbard, Signalling paediatric side effects using an ensemble of simple study designs, Drug Saf. 37 (3) (2014) 163–170. doi:10.1007/ s40264-014-0137-z.

S. A. Goldman, Limitations and strengths of spontaneous reports data, Clin. Ther. 20 (1998) C40–C44. doi:10.1016/S0149-2918(98)80007-6.

M. McGue, M. Osler, K. Christensen, Causal inference and observational research the utility of twins, Perspectives on Psychological Science 5 (5) (2010) 546–556. doi:10.1177/1745691610383511.

A. D. McMahon, Approaches to combat with confounding by indication in observational studies of intended drug effects, Pharmacoepidemiol. Drug Saf. 12 (7) (2003) 551–558. doi:10.1002/pds.883.

P. B. Ryan, D. Madigan, P. E. Stang, J. Marc Overhage, J. A. Racoosin, A. G. Hartzema, Empirical assessment of methods for risk identification in healthcare data: results from the experiments of the Observational Medical Outcomes Partnership, Stat. Med. 31 (30) (2012) 4401–4415. doi:10. 1002/sim.5620.

J. M. Reps, J. M. Garibaldi, U. Aickelin, D. Soria, J. Gibson, R. Hubbard, Comparison of algorithms that detect drug side effects using electronic healthcare databases, Soft. comput. 17 (12) (2013) 2381–2397. doi:10.1007/s00500-013-1097-4.

O. Caster, G. N. Nor´en, D. Madigan, A. Bate, Large-scale regression-based pattern discovery: The example of screening the WHO global drug safety database, Stat. Anal. Data. Min. 3 (4) (2010) 197–208. doi:10.1002/sam.10078.

R. Harpaz, K. Haerian, H. S. Chase, C. Friedman, Mining electronic health records for adverse drug effects using regression based methods, in: Proceedings of the 1st ACM International Health Informatics Symposium, ACM, 2010, pp. 100–107. doi:10.1145/1882992.1883008.

Y. Li, H. Salmasian, S. Vilar, H. Chase, C. Friedman, Y. Wei, A method for controlling complex confounding effects in the detection of adverse drug reactions using electronic health records, J. Am. Med. Inform. Assoc. 21 (2) (2014) 308–314. doi:10.1136/amiajnl-2013-001718.

J. Bien, J. Taylor, R. Tibshirani, et al., A lasso for hierarchical interactions, The Annals of Statistics 41 (3) (2013) 1111–1141. doi:10.1214/ 13-AOS1096.

J. Zhou, F. Wang, J. Hu, J. Ye, From micro to macro: data driven phenotyping by densification of longitudinal electronic medical records, in: Proceedings of the 20th ACM SIGKDD international conference on Knowledge discovery and data mining, ACM, 2014, pp. 135–144. doi:10.1145/2623330.2623711.

G. Dong, J. Li, Efficient mining of emerging patterns: Discovering trends and differences, in: Proceedings of the fifth ACM SIGKDD international

conference on Knowledge discovery and data mining, ACM, 1999, pp. 43–52. doi:10.1145/312129.312191.

G. J. Simon, J. Schrom, M. R. Castro, P. W. Li, P. J. Caraballo, Survival association rule mining towards type 2 diabetes risk assessment, in: AMIA Annual Symposium Proceedings, Vol. 2013, American Medical Informatics Association, 2013, pp. 1293–1302.

S. Schneeweiss, J. A. Rassen, R. J. Glynn, J. Avorn, H. Mogun, M. A. Brookhart, High-dimensional propensity score adjustment in studies of treatment effects using health care claims data, Epidemiology (Cambridge, Mass.) 20 (4) (2009) 512–522. doi:10.1097/EDE.0b013e3181a663cc.

M. A. Brookhart, S. Schneeweiss, K. J. Rothman, R. J. Glynn, J. Avorn, T. Stu¨rmer, Variable selection for propensity score models, Am. J. Epidemiol. 163 (12) (2006) 1149–1156. doi:10.1093/aje/kwj149.

J. M. Brooks, R. L. Ohsfeldt, Squeezing the balloon: propensity scores and unmeasured covariate balance, Health services research 48 (4) (2013) 1487–1507. doi:10.1111/1475-6773.12020.

J. D. Lewis, R. Schinnar, W. B. Bilker, X. Wang, B. L. Strom, Validation studies of the health improvement network (THIN) database for pharmacoepidemiology research, Pharmacoepidemiol. Drug Saf. 16 (4) (2007) 393–401. doi:10.1002/pds.1335.

S. L. Spruance, J. E. Reid, M. Grace, M. Samore, Hazard ratio in clinical trials, Antimicrobial agents and chemotherapy 48 (8) (2004) 2787–2792. doi:10.1128/AAC.48.8.2787-2792.2004.

D. R. Cox, Partial likelihood, Biometrika 62 (2) (1975) 269–276. doi:10.1093/biomet/62.2.269.

A. Hoerl, R. Kennard, Ridge regression, Encyclopedia of statistical sciences 8 (1988) 129–136. doi:10.1002/0471667196.ess2280.

R. Tibshirani, Regression shrinkage and selection via the lasso, J. R. Stat. Soc. Series B Stat. Methodol. 58 (1) (1996) 267–288.

H. Zou, T. Hastie, Regularization and variable selection via the elastic net, J. R. Stat. Soc. Series B Stat. Methodol. 67 (2) (2005) 301–320.

L. E. Chambless, G. Diao, Estimation of time-dependent area under the ROC curve for long-term risk prediction, Stat. Med. 25 (20) (2006) 3474–3486. doi:10.1002/sim.2299.

R Core Team, R: A Language and Environment for Statistical Computing, R Foundation for Statistical Computing, Vienna, Austria (2015).
URL https://www.R-project.org/

M. Hahsler, B. Grün, K. Hornik, A computational environment for mining association rules and frequent item sets, J. Stat. Softw. 14 (15) (2005) 1–15.

T. M. Therneau, P. M. Grambsch, Modeling survival data: extending the Cox model, Springer Science & Business Media, 2000.

N. Simon, J. Friedman, T. Hastie, R. Tibshirani, et al., Regularization paths for Coxs proportional hazards model via coordinate descent, J. Stat. Softw. 39 (5) (2011) 1–13.